%% file: fff-arxiv.tex
\documentclass{article}

\usepackage{PRIMEarxiv}

\usepackage[utf8]{inputenc} 
\usepackage[T1]{fontenc}    
\usepackage{hyperref}       
\usepackage{url}            
\usepackage{booktabs}       
\usepackage{amsfonts}       
\usepackage{nicefrac}       
\usepackage{microtype}      
\usepackage{lipsum}
\usepackage{fancyhdr}       
\usepackage{graphicx}       
\usepackage{amsmath}
\usepackage{amssymb}
\usepackage{booktabs}
\usepackage{subcaption}

\usepackage{algorithm}
\usepackage{algorithmic}
\graphicspath{{media/}}     

\newtheorem{definition}{Definition}

\newcommand{\methname}{FFF}
\newcommand{\iimeth}{implicitly improved method}
\newcommand{\aatype}{AA-type}
\newcommand{\angstrom}{\mbox{\normalfont\AA}}
\newcommand{\cref}[1]{\ref{#1}}
\newcommand{\ppvec}{pseudo-peptide vector}

\newcommand{\Calpha}{C$_\alpha$}

\begin{document}

\title{\methname{}: Fragments-Guided Flexible Fitting for Building Complete Protein Structures}

\author{Weijie Chen\textsuperscript{a,b}, Xinyan Wang\textsuperscript{a},
and Yuhang Wang\textsuperscript{a,}\thanks{Corresponding author}\\
\texttt{baoz@pku.edu.cn, wangxy940930@gmail.com, 
stevenwaura@gmail.com}\\
$^a$ DP Technology, Ltd., Beijing, China; 
$^b$ Peking University, Beijing, China
}

\maketitle

\input{0-abstract}

\input{1-introduction}

\input{2-related_work.tex}

\input{3-method.tex}

\input{4-experiment.tex}

\input{5-conclusion.tex}

{\small
\bibliographystyle{ieee_fullname}
\bibliography{dpemff_refs}
}

\end{document}

%% file: 0-abstract.tex
\begin{abstract}
Cryo-electron microscopy (cryo-EM) is a technique for reconstructing the 3-dimensional (3D) structure of biomolecules (especially large protein complexes and molecular assemblies). As the resolution increases to the near-atomic scale, building protein structures \textit{de novo} from cryo-EM maps becomes possible. Recently, recognition-based \textit{de novo} building methods have shown the potential to streamline this process. However, it cannot build a complete structure due to the low signal-to-noise ratio (SNR) problem. At the same time, AlphaFold has led to a great breakthrough in predicting protein structures. This has inspired us to combine fragment recognition and structure prediction methods to build a complete structure. In this paper, we propose a new method named \methname{} that bridges protein structure prediction and protein structure recognition with flexible fitting. First, a multi-level recognition network is used to capture various structural features from the input 3D cryo-EM map. 
Next, protein structural fragments are generated using pseudo peptide vectors and a protein sequence alignment method based on these extracted features.
Finally, a complete structural model is constructed using the predicted protein fragments via flexible fitting. Based on our benchmark tests, \methname{} outperforms the baseline methods for building complete protein structures.

\end{abstract}

%% file: 1-introduction.tex
\section{Introduction}
\label{sec:intro}
With the advances in hardware and image processing algorithms, cryo-EM has become a major experimental technique for determining the structures of biological macro-molecules, especially proteins. Cryo-EM data processing consists of two major steps:
3D reconstruction and structure building. In the 3D reconstruction step, 
a set of 2-dimensional (2D) micrographs (projection images) are collected using transmission electron microscopy for biological samples embedded in a thin layer of amorphous ice. Each micrograph contains many 2D projections of the molecules in unknown orientations. Software tools such as 
RELION~\cite{scheres2012relion}, 
cryoSPARC~\cite{punjani2017cryosparc} and 
cryoDRGN~\cite{zhong2021cryodrgn} 
can be used to recover the underlying 3D molecular density map.

In the second step, we build the atomic structure of the underlying protein by trying to determine  the positions of all the atoms using the 3D density map from the 3D reconstruction step. This is done through an iterative process that contains the following steps:
(1) rigid-body docking of an initial staring structure into the cryo-EM map;
(2) manual or automated/semi-automated flexible refinement of the docked structure to match the map
~\cite{wriggers1999situs,roseman2000docking,woetzel2011bcl}.
The manual process is extremely time consuming
and requires extensive expert-level domain knowledge.
Existing automated/semi-automated flexible fitting may mismatch
the initial structure and density map regions.
\begin{figure}[!t]
  \centering
    \includegraphics[scale=0.2]{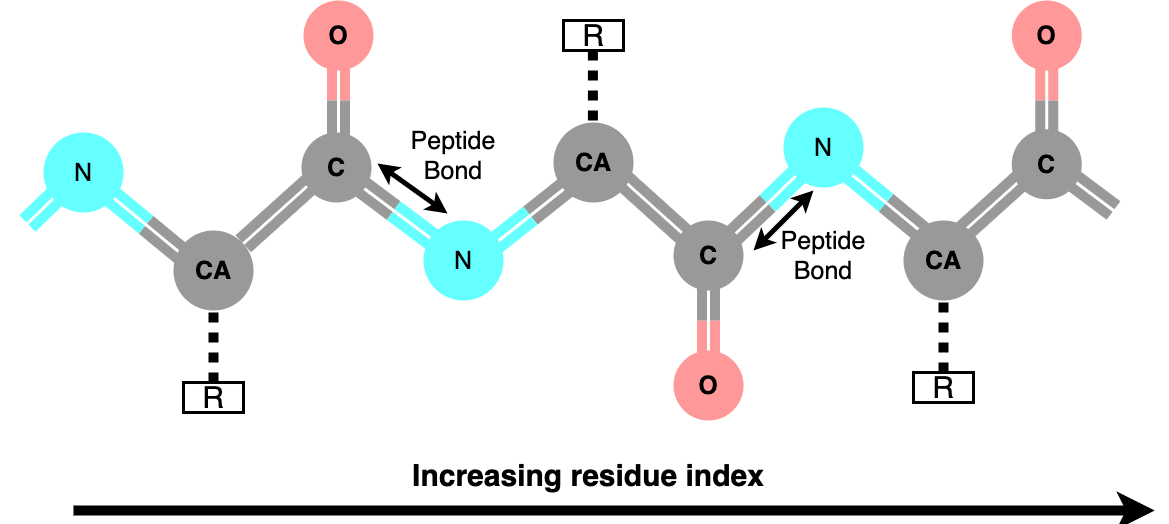}
    \caption{A protein backbone along the chain with increasing residue index from left to right. Arrow marks the peptide bond linking two consecutive amino acids from C atom of one amino acid and N atom of another.}
    \label{fig:protein}
    \vspace{-0.2cm}
\end{figure}

In the past decades, the resolution of cryo-EM has been drastically improved from medium-resolution (5--10~\AA{}) to near-atomic resolution (1.2--5~\AA{})~\cite{si2022artificial}.
At the same time, deep learning-based \textit{de novo} protein structure building methods have shown great progress. As a result, building a reasonable atomic structural model \textit{de novo} is already feasible for maps whose resolutions are better than 3~\AA{}. Even so, due to the flexible nature of some local structures, \textit{de novo} modeling methods often cannot model a complete protein structure and still require significant manual effort and domain expertise.

With the advent of AlphaFold~\cite{jumper2021highly}, high-accuracy structure predictions have greatly helped biologists in structure modeling from cryo-EM maps. AlphaFold can predict
single-chain structures that closely match the cryo-EM density maps in many cases~\cite{akdel2022structural}.
However, its ability is still limited to predicting
structures for
multimeric protein complexes, a protein with alternative conformations,
and protein-ligand complexes.
Cryo-EM studies are required in solving structures in these complex scenarios.

One possible solution is to effectively combine experimental information and machine learning-based prediction. In this work, we propose a new method called \methname{} (``Fragment-guided Flexible Fitting") that enables more reliable and complete cryo-EM structure building by bridging protein structure prediction and protein structure recognition with flexible fitting. First, we use a multi-level recognition network to capture different-level structural features from the input 3D volume (i.e. cryo-EM map).
Next, the extracted features are used for fragment recognition to construct the recognized structure. Finally, we run a flexible fitting to building a complete structure based on the recognized structure.

Our main contributions are as follows:
\begin{itemize}
  \item We propose a more straightforward method for protein backbone tracing using pseudo peptide vectors.
  \item We propose a more sensitive and accurate protein sequence alignment algorithm to identify the amino acid types and residue index of detected residues. This is essential for aligning the recognized fragments and the machine-learning predicted structure.
  \item We combine deep-learning-based 3D map recognition with molecular dynamics that surpasses previous methods in cryo-EM structure building using AlphaFold.
\end{itemize}

%% file: 2-related_work.tex
\section{Related Work}
\label{sec:relwork}
\subsection{Fragments Recognition in \textit{de novo} Building}
Object recognition methods based on deep learning have been widely used in 2D images, videos, and multi-camera 3D scenes. With the revolution in cryo-EM resolution and the development of cryo-EM databases, deep learning methods began to be applied to protein structure recognition in \textit{de novo} building from cryo-EM maps.

$A^{2}$Net~\cite{xu2019a2} proposed a two-stage detection workflow for identifying protein structures from cryo-EM maps. In the first stage, $A^{2}$Net use region proposal network (RPN)~\cite{ren2015faster} to generate proposals of amino acid locations. In the second stage, with the amino acid proposals, they extract the region of interest and predict its amino acid type and coordinates through 3D convolution neural networks (CNN). With the amino acid proposals, they construct a Monte Carlo Tree Search (MCTS) for backbone threading.

Recently, DeepTracer~\cite{pfab2021deeptracer}, as a fast and fully automated deep learning method for modeling atomic structure from cryo-EM maps, has shown outstanding performance compared with the existing methods. DeepTracer contains four different U-Nets~\cite{ronneberger2015u} for protein recognition and a heuristic traveling salesman problem (TSP) algorithm for constructing an atomic protein structure.

Compared with the previous \textit{de novo} methods, 
\methname{} has a simpler and effective design for atom recognition,
which is a extremely difficult task for most cryo-EM maps. We focus on the detectable fragments (often with stable secondary structures, like $\alpha$-helix~\cite{si2012machine}) and designed a network architecturefor one-stage detection. For protein backbone tracing, 
residues are directly connected using predicted pseudo peptide vectors 
instead of relying on heuristic searching algorithms (see Methods).
Most importantly, we propose a more sensitive and accurate protein sequence alignment algorithm that helps us to match the predicted fragments with the starting structure.

\subsection{Structure Refinement through Flexible Fitting}
There are two popular flexible fitting methods for cryo-EM
structure refinement: Molecular Dynamics flexible Fitting (MDFF)
~\cite{Trabuco2008,Trabuco2009}
and Correlation-Driven Molecular Dynamics (CDMD)~\cite{Igaev2019}.
In MDFF, the experimental cryo-EM density map is converted into a potential
energy function $U_{EM}^{\text{MDFF}}$ defined as below.
\begin{equation}
  U_{EM}^{\text{MDFF}} = k \cdot \left(1 - \frac{\rho(\vec{r})}{\rho_{max}}\right)
  \label{eqn:mdff-U}
  \end{equation}
where $\rho(\vec{r})$ is the cryo-EM density value at cartesian coordinate $\vec{r}$, the normalization factor $\rho_{max}=\max_{\vec{r}}\rho(\vec{r})$. $k$ is a global parameter for adjusting the density-derived forces.

With this formulation, atoms of the initial structure are driven towards the energy minima defined by $U_{EM}^{\text{MDFF}}$ which correspond to locations of target atoms in the cryo-EM experiments.

In CDMD, the initial structure is biased toward the cryo-EM density map
through an EM potential defined by the correlation-correlation
coefficient (ccc) between the experimental map and a simulated map based
on the conformations (a set of atomic coordinates) sampled during the MD
simulations.
\begin{equation}
    U_{EM}^{\text{CDMD}} = k \cdot \left(
        1 -
        \frac{\sum_{i=1}^N \rho_{exp}(\vec{r}_i)\cdot \rho_{sim}(\vec{r}_i)}{\sqrt{\sum_{i=1}^N \rho_{exp}^2(\vec{r}_i)   \cdot
            \sum_{i=1}^N \rho_{sim}^2(\vec{r}_i) }}
        \right)
\end{equation}
The fitting terminates when the ccc reaches the target threshold.

Both MDFF and CDMD can iteratively refine cryo-EM structures.
The accuracy of both methods are limited by the lack of the correspondence
information between atoms in the initial structure and regions in the input cryo-EM map. If the initial structure deviates significantly from
the cryo-EM map (low ccc value), the flexible fitting may fail.
\methname{} is designed to overcome this limitation by adding
atomic-level constraints constructed based on  cryo-EM map recognition.

\subsection{Cryo-EM Building meets AlphaFold}
With breakthrough methods in \textit{de novo} protein structure prediction such as AlphaFold~\cite{jumper2021highly}, some structural biologists try to fit the predicted structures from AlphaFold into cryo-EM maps through rigid-body alignment~\cite{he2022model}. However, for poorly predicted structures, this approach often requires expert knowledge for manual adjustments.

Using custom structural templates is considered a potentially effective technique for enabling AlphaFold to predict alternative conformations of the same protein~\cite{jumper2021applying}. Recent work attempted to fit the \textit{de novo} predicted structure into the cryo-EM map and then applied the refined structure as custom template features to implicitly guide the structure prediction process of AlphaFold (\iimeth{})~\cite{terwilliger2022improved}. 
However, these user-supplied template features may be misleading or ignored. For these cases, AlphaFold cannot predict desired structures by simply replacing template features.

%% file: 3-method.tex
\section{Method}
\label{sec:method}
\begin{figure*}[!t]
  \centering
    \includegraphics[scale=0.4]{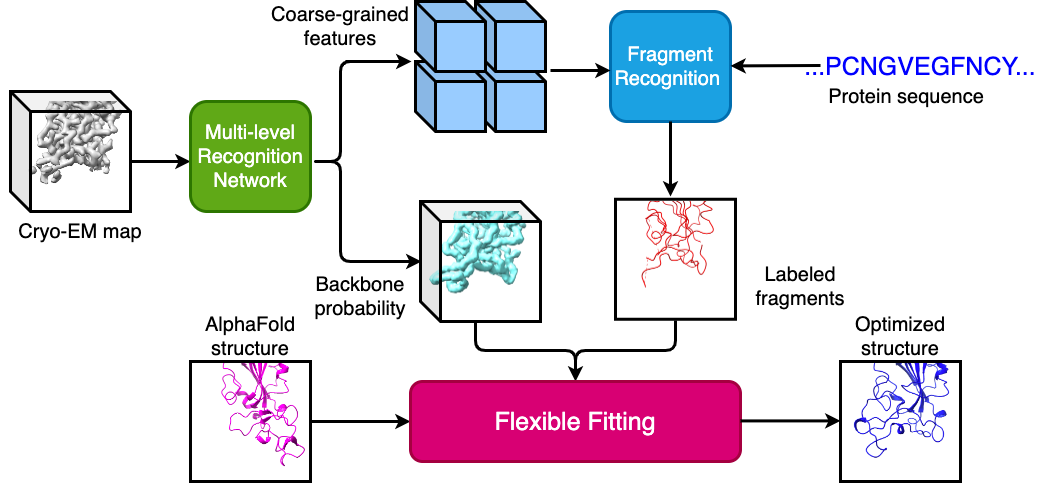}
    \caption{The workflow of \methname{}: the multi-level recognition network extracts coarse-grained features and predicts backbone probability map from the input cryo-EM map. The given protein sequence and coarse-grained features are utilized to convert structural information into labeled fragments. With the guidance of backbone probability map and  labeled fragments, the initial structure will be optimized by the flexible fitting.}
    \label{fig:pipeline}
\end{figure*}
Our method consists of three steps (see ~\cref{fig:pipeline}). The first step contains a multi-level deep neural network for one-stage residue recognition, which predicts coarse-grained features of C$\alpha$ atoms in each residue and a backbone probability map. In the second step, we connect the C$\alpha$ atoms of residues to fragments with neighbor selecti  on based on estimated \ppvec{}s from the first stage and annotated using the given protein sequence. In the final step, we run molecular dynamics simulations with the constraints from predicted protein fragments and the backbone map to optimize the initial AlphaFold prediction.

\subsection{Training settings}
We collected around 2400 cryo-EM maps whose resolution range from 1 \AA{}  to 4 \AA{} from the EM Data Bank (EMDB), and the corresponding modeled PDB files that were released before 2020-05-01 from Protein Data Bank (PDB). The release date is consistent with the training setting of AlphaFold to prevent data leakage. The following cases are filtered from the training data:
\begin{itemize}
  \item cryo-EM maps that are misaligned with the deposited structures
  \item cryo-EM maps with large regions where  modeled structures are missing
  \item modeled structures that do not have any corresponding cryo-EM map density
  \item duplicated structures
  \item maps with other types of macromolecules, e.g. nucleic acids
\end{itemize}
To standardize the input maps, we resampled the cryo-EM maps with a voxel size of $1\times1\times1\,\angstrom^{3}$. To prepare training data for the map-recognition network, we cropped the original cryo-EM maps into small cubic sample regions of shape $(32,32,32)$. To balance the diversity of training samples and training efficiency, 80\% of the training data were restricted to only samples from map regions with protein structures and 20\% from samples without restrictions.

\subsection{Multi-level Recognition Network}
\label{subsec:network}
\begin{figure*}[!t]
    \includegraphics[scale=0.3]{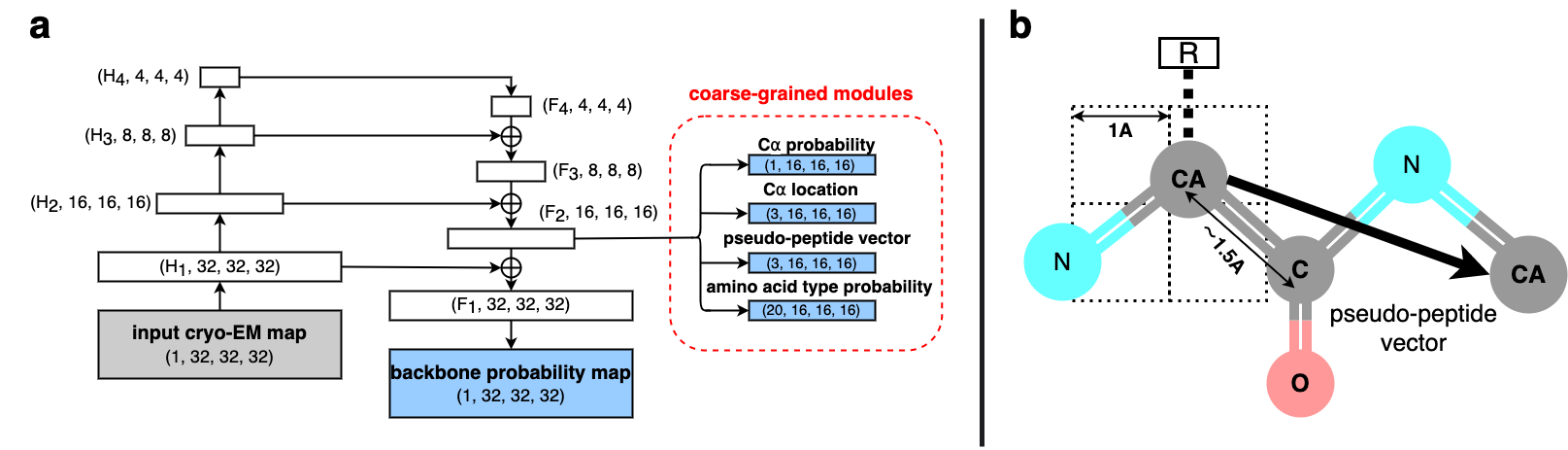}
    \caption{\textbf{(a)} the architecture of 3D RetinaNet and separate coarse-grained modules. \textbf{(b)} \ppvec{} between two consecutive C$\alpha$ atoms. The dot-line grid show the vision comparison between downsampling grid cell and tiny voxel.}
    \label{fig:network-ppv}
\end{figure*}
The architecture diagram of a multi-level recognition network is illustrated in~\cref{fig:network-ppv}a. Inspired by the RetinaNet architecture~\cite{lin2017focal}, we substituted all 2D convolutions to 3D convolutions and modified them to improve the performance for this task. The network not only predicts the voxel-wise probabilities of the backbone (BB) but also unifies four separate coarse-grained modules: C$\alpha$ detection, C$\alpha$ location prediction, \ppvec{} (PPV) estimation , and amino acid (AA) classification.

The BB part of the multi-level network determines whether each voxel belongs to the backbone or not. Thus, it's a 3D segmentation problem. To capture the global information and deal with highly imbalanced data, we use \textbf{Dice Loss}~\cite{milletari2016v} to measure the difference between predicted probabilities $y_{BB}$ and true labels $\hat{y}_{BB}$. The backbone loss is defined as follows:
\begin{equation}
\begin{aligned}
    \mathcal{L}_{BB} &= \text{DiceLoss}(y_{BB}, \hat{y}_{BB}) \\
    \text{DiceLoss}(X,Y) &= 1-\frac{2\|X \circ Y\|}{\|X^{2}\| + \|Y^{2}\|}
    \label{eq:BBloss}
\end{aligned}
\end{equation}
where $\|X\|=\sum_{i,j,k}X_{ijk}$, i.e. the sum of all elements in the tensor, $X \circ Y$ means the element-wise product of two tensors.

Our coarse-grained modules predict various features which correspond to the map with a grid size of  $2\times2\times2\, \angstrom^{3}$, instead of the original grid size of $1\times1\times1 \,\angstrom^{3}$ (\cref{fig:network-ppv}a).
This strategy increases the robustness of our method, especially for noisy and fragmented cryo-EM maps.
In addition, considering the ideal distance of consecutive C$\alpha$ atoms is about 3.8 \AA{} ~\cite{chakraborty2013protein}, a grid size of 2 \AA{} ensures at most one C$\alpha$ atom in any grid cell.

The C$\alpha$ detection module predicts the probability of whether a grid cell contains a C$\alpha$ atom. Considering the class imbalance and grid-wise precision, we combined Dice Loss and weighted Binary Cross Entropy (BCE) to leverage their benefits.
\begin{equation}
\begin{aligned}
    \mathcal{L}_{\text{rec}} =& \text{DiceLoss}(y_{C\alpha}, \hat{y}_{C\alpha}) + \text{BCE}(y_{C\alpha}, \hat{y}_{C\alpha}) \\
    \text{BCE}(X,Y) =& -\frac{1}{S^3}\times\sum_{i,j,k}^{S^{3}}
     \beta Y_{ijk}\log(X_{ijk}) + (1-\beta)(1-Y_{ijk})\log(1-X_{ijk})
    \label{eq:CaRecloss}
\end{aligned}
\end{equation}
where $y_{C\alpha}$, $\hat{y}_{C\alpha}$ represent the predicted probabilities and true labels respectively, $S^{3}$ is the number of grid cells. $\beta = 1-\|\hat{y}_{C\alpha}\|/S^3$ denoted the class-balanced weight.

The C$\alpha$ location module predicts the relative coordinates of C$\alpha$. The predicted offsets $\mathbf{X}_{C\alpha}=(x,y,z)\in [0,2]^{3}$ represent the location of C$\alpha$ atom relative to the boundaries of the grid cell. We defined the mean square error of location among the grid cells containing C$\alpha$ atom:
\begin{equation}
  \begin{aligned}
      \mathcal{L}_{\text{loc}} &= \frac{1}{N_{C\alpha}}\sum_{i}^{N_{C\alpha}}\|\mathbf{X}_{C\alpha^{i}}-\hat{\mathbf{X}}_{C\alpha^{i}}\|_2^{2}
      \label{eq:CaLocloss}
  \end{aligned}
\end{equation}
where $N_{C\alpha}$ is the number of grid cells that contains \Calpha{} atoms and $\hat{\mathbf{X}}_{C\alpha^{i}}$ is the true label.

Generally, there are 20 types of amino acids that make up the proteins found in nature. The amino acid classification module predicts the conditional amino acid type (\aatype{}) probabilities $P(AA|C\alpha)$ of each grid cell. These probabilities are conditioned on the grid cell containing a C$\alpha$ atom (i.e. the central carbon atom of a single amino acid residue). Thus, we can compute the average of cross entropy within $N_{C\alpha}$ grid cells:
\begin{equation}
  \begin{aligned}
      \mathcal{L}_{AA} &= -\frac{1}{N_{C\alpha}}\sum_{i}^{N_{C\alpha}}\sum_{j}^{20}Z_{ij}\log P(AA_j|C\alpha^{i})
      \label{eq:AAloss}
  \end{aligned}
\end{equation}
where $Z_{ij}=1$ if $i$-th C$\alpha$ atom belongs to $j$-th \aatype{}, otherwise to be $0$.

Finally, the PPV estimation module predicts the \ppvec{}s pointing from the current grid's C$\alpha$ atom to its consecutive C$\alpha$ atoms (\cref{fig:network-ppv}b). Since a typical consecutive C$\alpha$–C$\alpha$ distance is about 3.8 \AA{}, we scale the standard hyperbolic tangent from $[-1,1]^{3}$ to $[-4.0, 4.0]^{3}$ and calculate the predicted \ppvec{}s. Similar to C$\alpha$ location, we compute the mean square error of vectors as follows:
\begin{equation}
  \begin{aligned}
      \mathcal{L}_{PPV} &= \frac{1}{N_{C\alpha}}\sum_{i}^{N_{C\alpha}}\|\mathbf{V}_{C\alpha^{i}}-\hat{\mathbf{V}}_{C\alpha^{i}}\|_2^{2}
      \label{eq:PPloss}
  \end{aligned}
\end{equation}
where $\mathbf{V}_{C\alpha^{i}}=(v_x,v_y,v_z)$ denotes the PPV of $i$-th C$\alpha$ atom.

The total loss function is defined as:
\begin{equation}
    \mathcal{L} = \mathcal{L}_{BB} + \lambda_{r}\mathcal{L}_{rec} + \lambda_{l}\mathcal{L}_{loc} + \lambda_a \mathcal{L}_{AA} +\lambda_p\mathcal{L}_{PPV}
\end{equation}
During training, we set $\lambda_r=\lambda_l=\lambda_a=1.0$ and $\lambda_p$=0.05.
\subsection{Fragment Recognition}
In related methods, a modified version of the TSP or MCTS algorithm is often used for connecting the detected C$\alpha$ atom into a chain (i.e., \Calpha{} tracing). However, on the one hand, false positive or false negative cases are unavoidable due to noise, non-protein molecules, or lost signals in experimental maps. These cases will result in some global or local errors when connecting $C_\alpha$ atoms. On the other hand, the factorial growth of the complexity makes it infeasible to find the optimum solution to connect these atoms into a chain in polynomial time.

To minimize the impact of false positives and keep the algorithmic complexity under control, we designed the PPV estimation module and formulate the following \Calpha{} tracing criterion:
\begin{definition}
  Any two C$\alpha$ atoms $q$ and $p$ are consecutive, if
  \begin{equation}
  \begin{aligned}
\|\tilde{\mathbf{X}}_q+\mathbf{V}_q - \tilde{\mathbf{X}}_{p}\|_2^{2}<= \epsilon
\end{aligned}
\end{equation}
\end{definition}
where $\tilde{\mathbf{X}}$ is the global coordinate of a \Calpha{} atom calculated from the predicted offsets $\mathbf{X}_{C\alpha}$ and grid cell indices; $\mathbf{V}$ is the \ppvec{} (see \cref{subsec:network}); $\epsilon$ is the threshold parameter.
\begin{figure}[!t]
  \centering
    \includegraphics[scale=0.4]{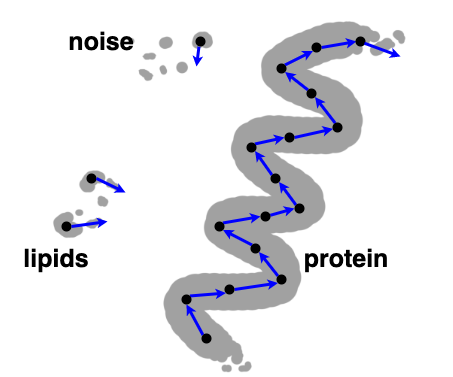}
    \caption{A simple instance illustrates how to connect detected C$\alpha$ atom  with the pseudo peptide vectors and then prune some false positive candidates recognized from lipids or noise.}
    \label{fig:pp}
    \vspace{-0.2cm}
\end{figure}

At inference time, we collected the indices of all C$\alpha$ candidates in the probability map above a given threshold and computed the necessary features from the corresponding predictions. Using the \Calpha{} tracing criterion defined above, one can connect the C$\alpha$ candidates to form continuous fragments. 
In practice, predicted PPV from false-positive signals (noise and non-protein small molecules) are unreliable and often result in very short predicted fragments.
Therefore, false positives can be effectively excluded by pruning away some short fragments and isolated atoms.

For a predicted fragment $D$ with $N$ residues, we can form a \aatype{} probability distribution matrix $P(AA|D)\in [0,1]^{N\times20}$ of fragment $D$
consisting of the predicted \aatype{} probabilities of the fragment residues. 
The full input protein sequence can be represented as a one-hot matrix F of shape $L\times20$, where $L$ is the total sequence length.
Its sub-sequence of length $N$ for residues $i$ to $i+N$ can be denoted by a sub-matrix of $F$, i.e., $F[i:i+N]\in \{0, 1\}^{N\times20}$.
Then, we can define an entropy-like score for measuring how well 
a sub-sequence matches the predicted \aatype{} probability distribution
of the fragment.
\begin{equation}
  s(i) = \frac{1}{N}\|\log(P(AA|D)) \circ F[i:i+N]\|
\end{equation}
The \aatype{}s and residue indexes of the fragment can be labeled by the sub-sequence with the highest score (i.e. the lowest entropy).
To measure the confidence of such alignment, we standardize the score to evaluate its significant degree against other alignments in the same fragment.
\begin{equation}
  \text{confidence}(D) = \frac{\max(s)-\text{mean}(s)}{\text{standard\_deviation}(s)+10^{-6}}
\end{equation}

\subsection{Building a Complete Protein Structure}
After the protein fragments and backbone map have been recognized, the complete
protein structure of the target protein is modeled in three steps.
These steps can be carried out successively or simultaneously.
Starting from an initial structure,
the first step updates this structure to match the recognized
fragments using targeted molecular dynamics (TMD)
~\cite{Schlitter1994}.
Compared to the conventional MD, TMD adds potential energy
term $U_{\text{TMD}}$ as defined below.

\begin{equation}
  U_{\text{TMD}} = \frac{1}{2}\, h \, \left(D(t)- \gamma(t) \cdot D(0)\right)^2
  \label{}
\end{equation}
\begin{equation}
  D(t) = \sqrt{\sum_{i=1}^N (\vec{r}_i(t) - \vec{c}_i)^2}
\end{equation}
where $h$ is the harmonic force constant for tuning the strength of the TMD forces.
$D(t)$ is a collective variable measuring the
minimum root-mean-square distance
(RMSD) between the atomic coordinate set $\{\vec{r}_i(t)\}_{i=1...N}$
sampled at time $t$ in TMD and
the reference coordinate set $\{\vec{c}_i\}_{i=1...N}$ from recognized fragments.
$N$ is the total number of atoms
where the TMD forces are applied.
$\gamma(t)$ is a scaling constant that changes from 1 to 0
as $t$ increases to the target simulation time $t_{total}$,
so that the RMSD value gradually decreases to zero.

The second step updates the complete backbone conformation to match
the predicted backbone map using
molecular dynamics flexible fitting (MDFF)
~\cite{Trabuco2008,Trabuco2009}.
During MDFF,
positional restraints are added to the atoms selected in the TMD step
to prevent large deviations.
Optionally, an additional MDFF can be performed with the experimental cryo-EM map to refine the orientation of the side-chain, meanwhile, backbone atoms are restrained to their positions.

%% file: 4-experiment.tex
\section{Experiments}
\label{sec:exp}
\subsection{Case Study: Alternative Conformations}
\begin{figure*}[!t]
\centering
  \begin{subfigure}{0.3\linewidth}
    \includegraphics[scale=0.16]{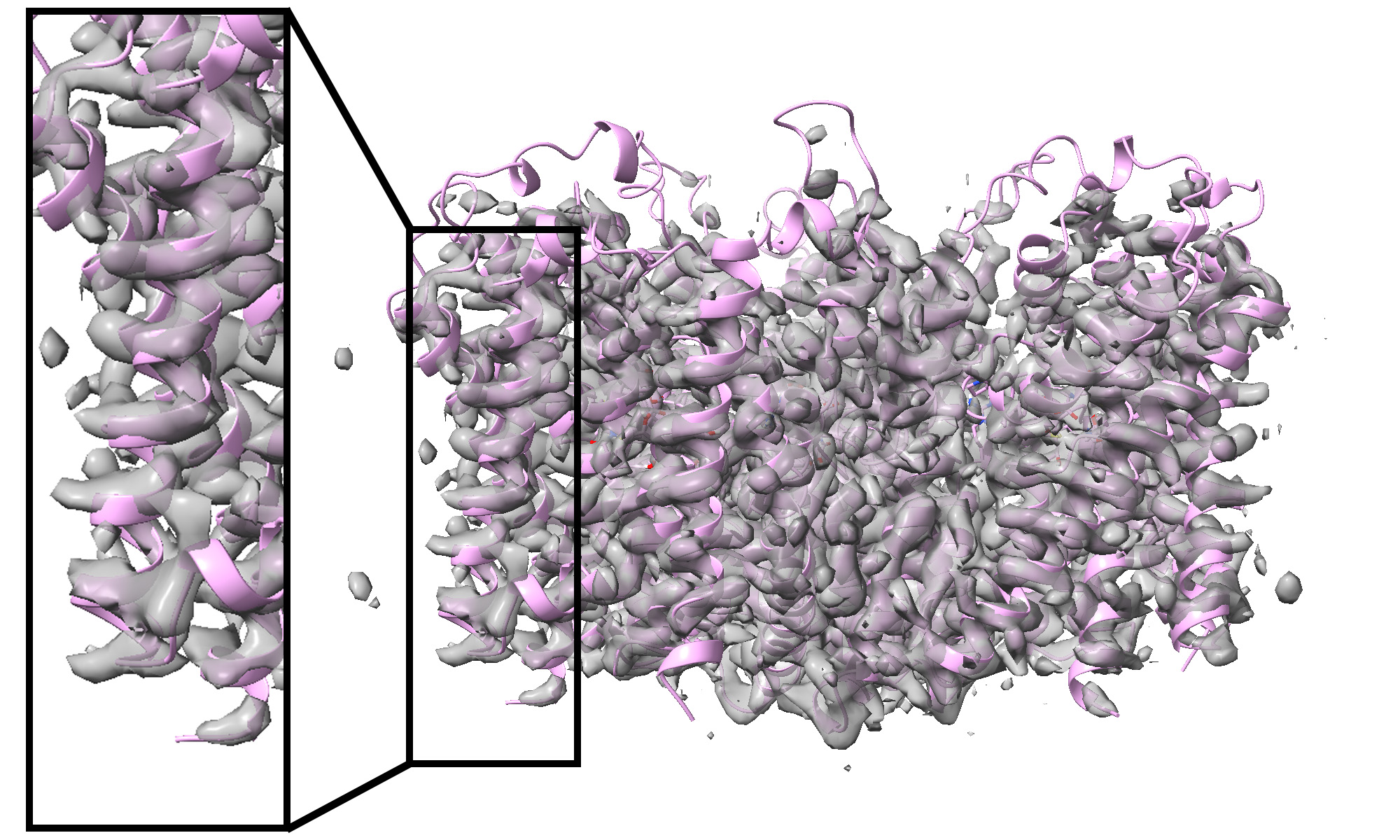}
    \caption{}
    \label{fig:outward}
  \end{subfigure}
  \hspace{2cm}
  \begin{subfigure}{0.3\linewidth}
    \includegraphics[scale=0.16]{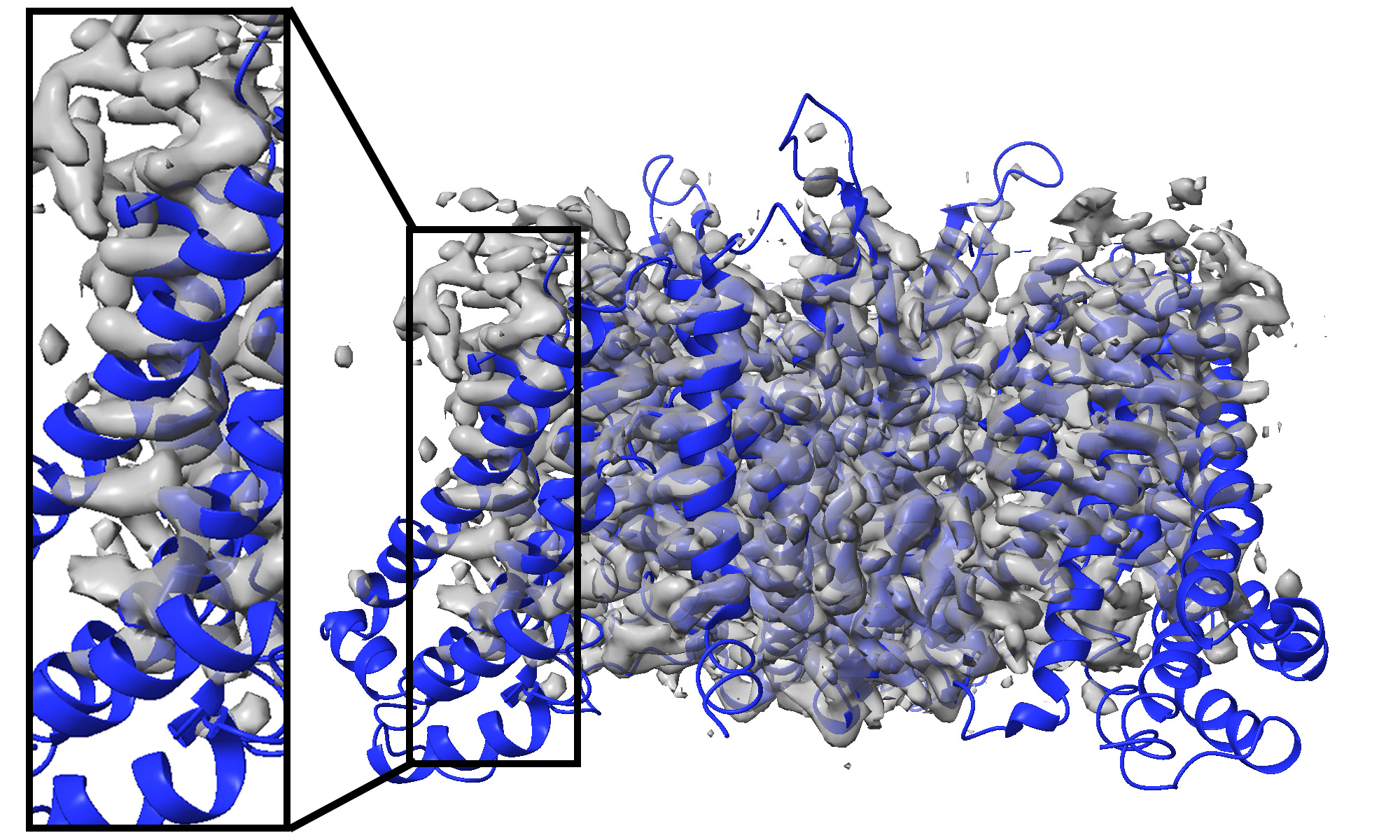}
    \caption{}
    \label{fig:inward}
  \end{subfigure}
  \vfill
  \begin{subfigure}{0.3\linewidth}
    \includegraphics[scale=0.16]{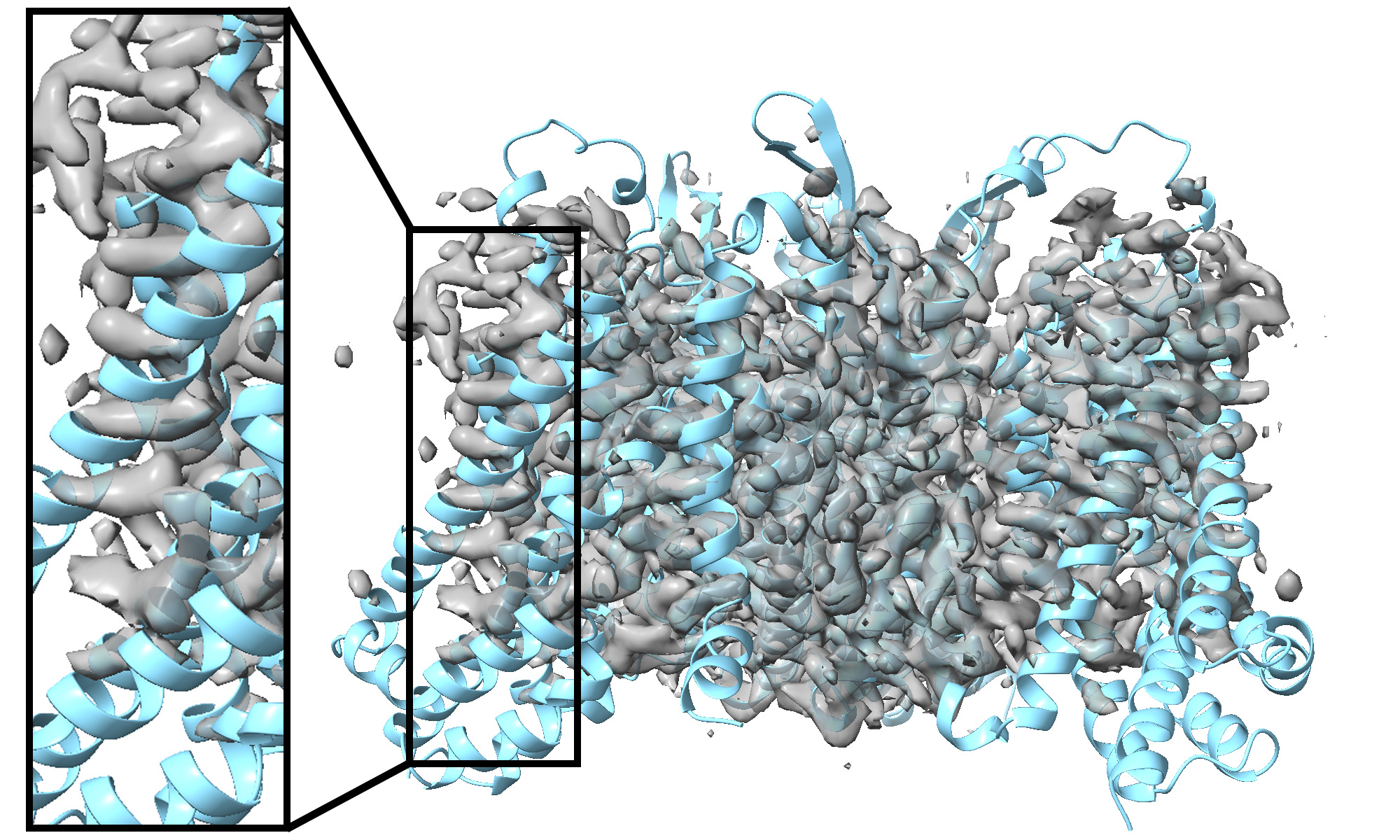}
    \caption{}
    \label{fig:af2}
  \end{subfigure}
  \hspace{0.2cm}
  \begin{subfigure}{0.3\linewidth}
    \includegraphics[scale=0.16]{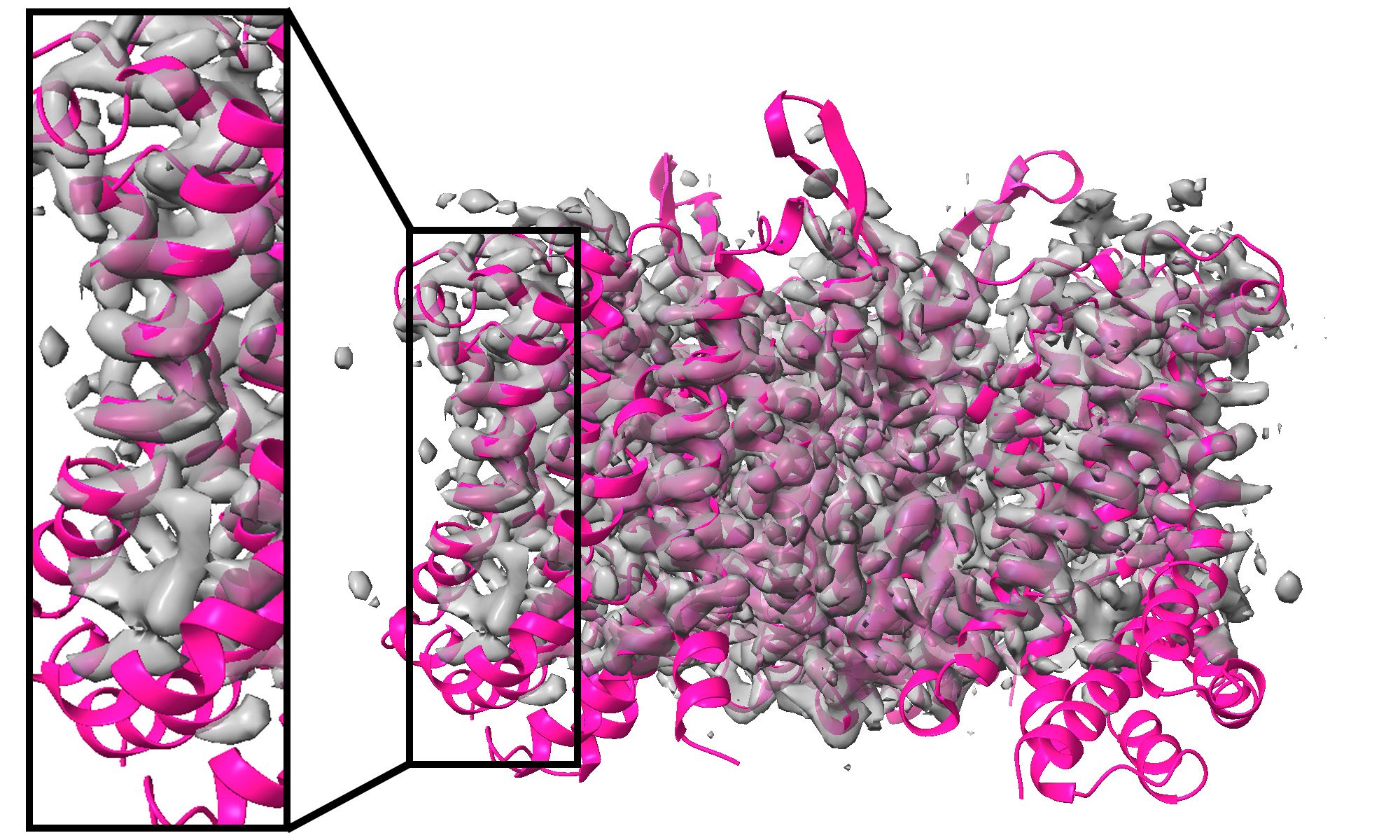}
    \caption{}
    \label{fig:mdff}
  \end{subfigure}
  \hspace{0.2cm}
  \begin{subfigure}{0.3\linewidth}
    \includegraphics[scale=0.16]{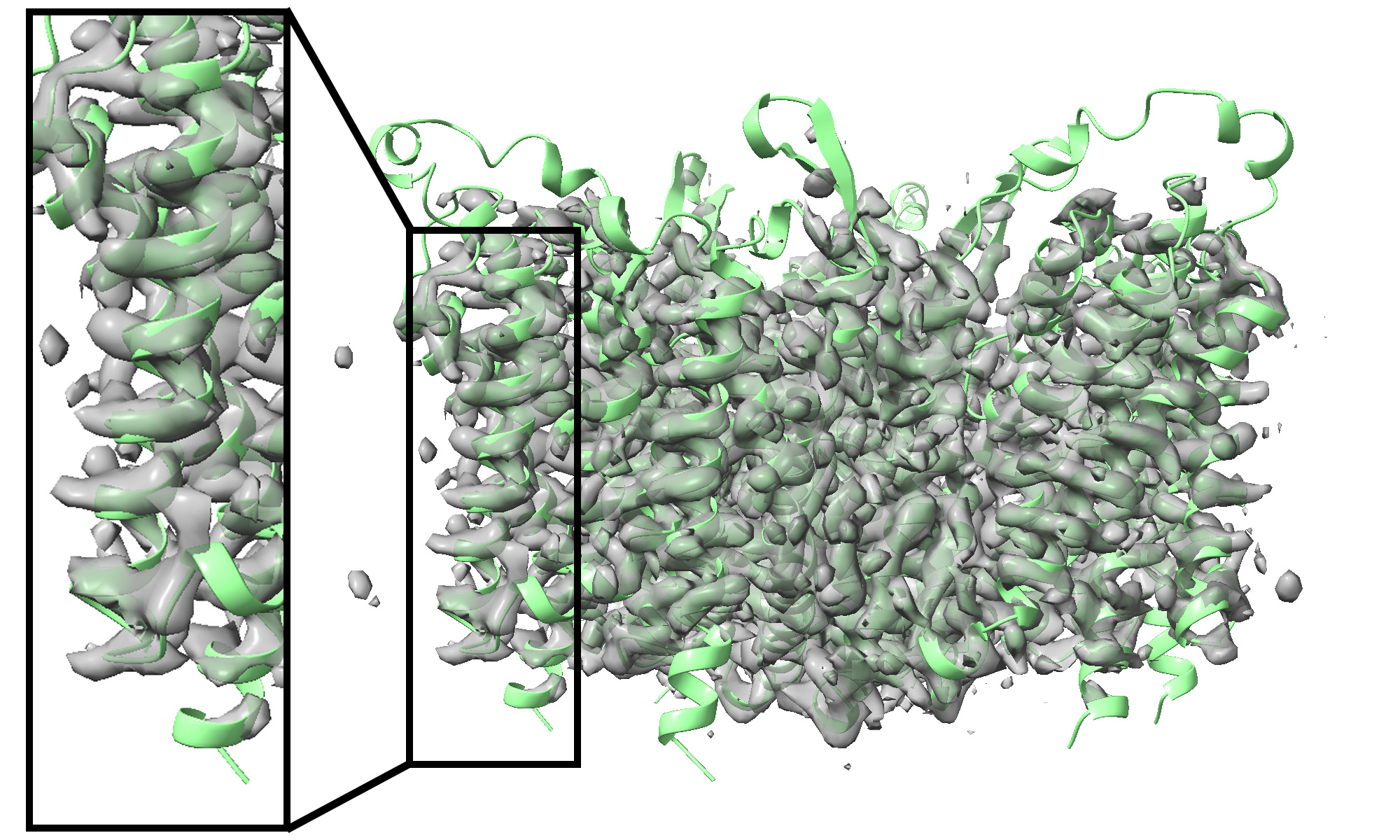}
    \caption{}
    \label{fig:our}
  \end{subfigure}
    \caption{Dual conformation of ASCT2 with the cryo-EM map of outward-open conformation. (a) shows the outward-open conformation. (b) shows the inward-open conformation that obviously mismatches with the map. (c) shows the prediction from AlphaFold. (d) shows the optimized structure with traditional flexible fitting. (e) shows the optimized structure with our method \methname{}}
    \label{fig:ASCT2}
\end{figure*}
AlphaFold sometimes cannot predict protein structures with matching conformation for proteins with multiple alternative stable conformations.

ASCT2 (SLC1A5) is a sodium-dependent neutral amino acid transporter that controls amino acid homeostasis in peripheral tissues~\cite{scalise2018human}. The molecular structure explains the alternating-access mechanism in which the transporter alternates between the outward-facing and inward-facing conformations (see \cref{fig:ASCT2}). Specifically, the inward-open conformation~\cite{garaeva2019one} of ASCT2 (PDB entry: 6RVX, released data: 2019-08-07) was solved first. Subsequently, the outward-open conformation~\cite{garibsingh2021rational} (PDB entry: 7BCQ, released data: 	2021-09-22) was solved from the corresponding map (EMDB entry: EMD-12142, resolution: 3.4 \AA{}), but excluded from the training data of AlphaFold.

From the perspective of structure building, 
many local regions in the ASCT2 cryo-EM maps have low SNR, making it difficult to automatically build
the corresponding local atomic structures \textit{de novo}.

On the other hand, the standard AlphaFold pipeline predicts the inward-open conformation (see \cref{fig:af2}) due to biased training data. Using traditional flexible fitting methods, the structure cannot be well fit into the map because of the significant differences between the initial and target structures (see \cref{fig:mdff}).

In comparison, our method captured more structural information from the map with a multi-level recognition network and explicitly guided the fitting of the initial structure into the map using predicted protein fragments (\cref{fig:our}). 
With this approach, we can effectively integrate experimental data into building a complete protein structure of a specific conformation.

\subsection{Performance Benchmark}
To validate the performance of our method, we collect the 24 protein structures excluded from the train data of AlphaFold. The resolution of cryo-EM maps ranges from 2.4 $\angstrom$ to 4.2 $\angstrom$.
To keep the assessment of the results simple,
only the map region for a single chain was kept for each
experimental cryo-EM map.

We compared results from AlphaFold (AF), traditional flexible fitting (AF+MDFF), implicit AlphaFold method (implicit AF)~\cite{terwilliger2022improved}, and \methname{} in terms of structural deviations from the deposited structures. TM-score~\cite{zhang2004scoring} was chosen as the metric for assessing the global structural similarities between protein structures.
We divided the benchmark dataset into two groups based on the quality of AlphaFold predictions.
The ``better group" (14 cases) has more accurate structure predictions (TM-score $\geq0.9$),
while the quality of AlphaFold predictions in the ``worse group" (10 cases) is not as good (see \cref{tab:benchmark}).
During the evaluation of our method, map fragments were pruned by removing predicted fragments with shorter than three residues and low confidence scores ($<3.4$) to avoid mismatching of protein fragments at later steps.

\begin{figure}[!t]
  \begin{subfigure}{0.23\linewidth}
    \includegraphics[scale=0.18]{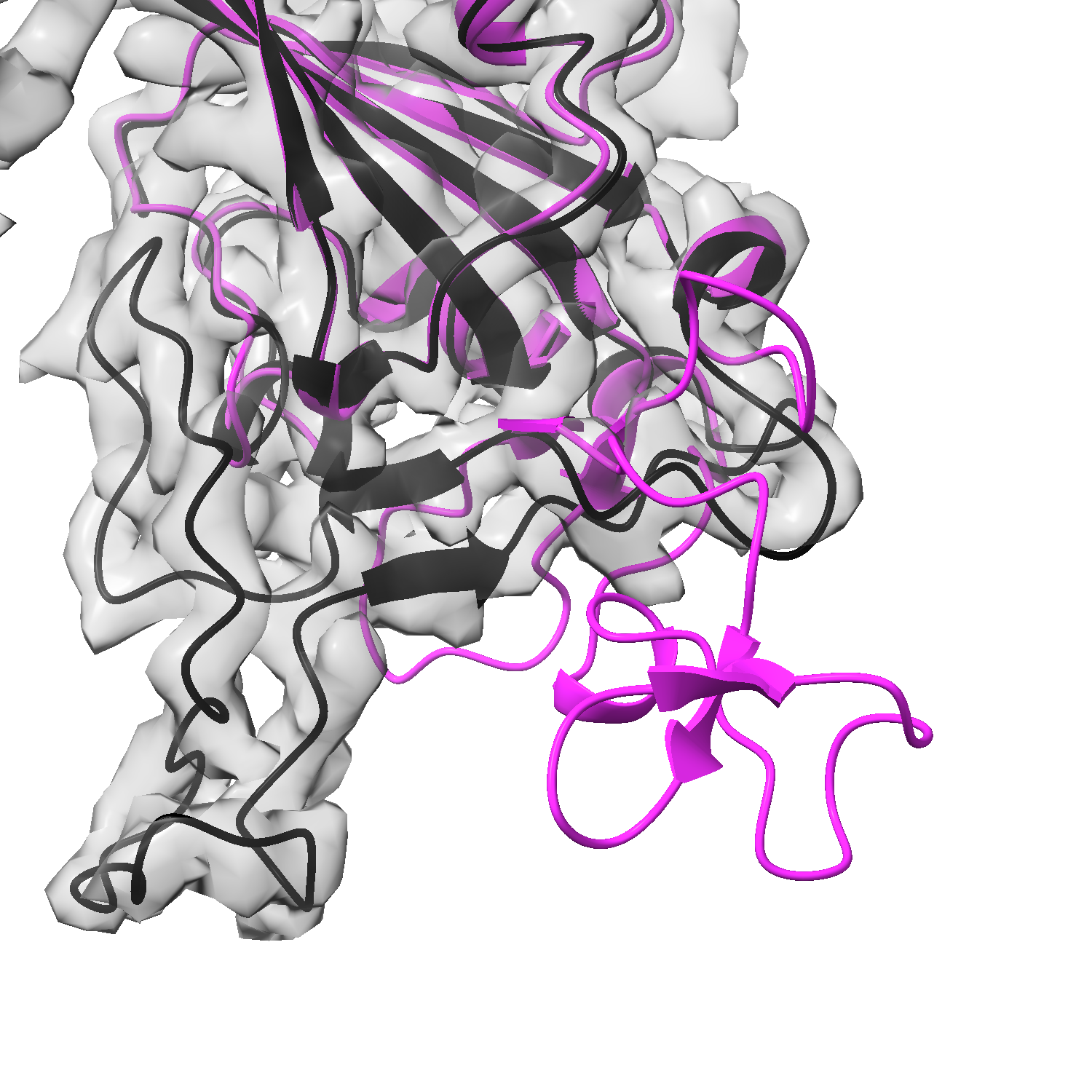}
    \caption{}
    \label{fig:af2-vs-true}
  \end{subfigure}
  \hspace{0.3cm}
  \begin{subfigure}{0.23\linewidth}
    \includegraphics[scale=0.18]{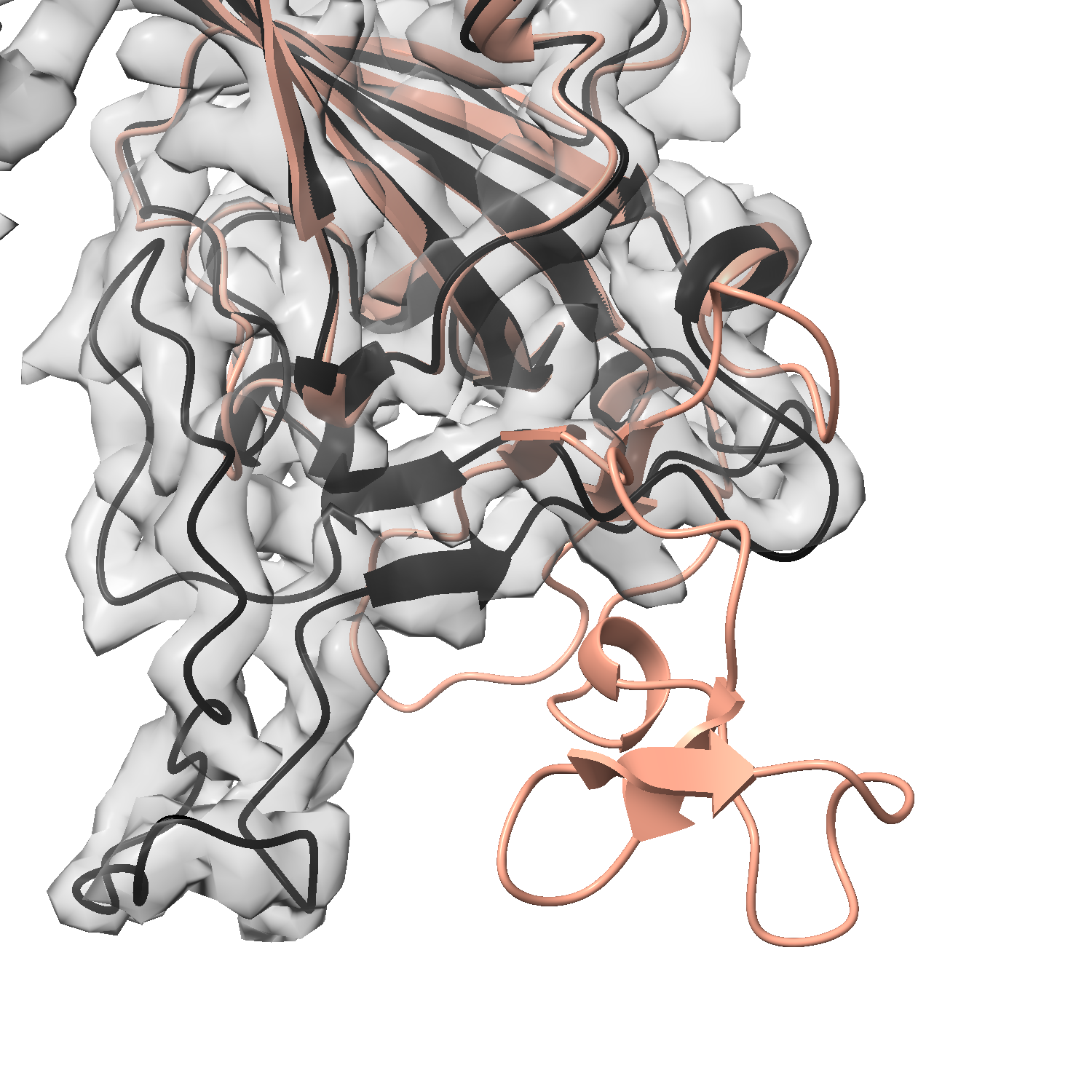}
    \caption{}
    \label{fig:mdff-vs-true}
  \end{subfigure}
  \hspace{0.3cm}
  \begin{subfigure}{0.23\linewidth}
    \includegraphics[scale=0.18]{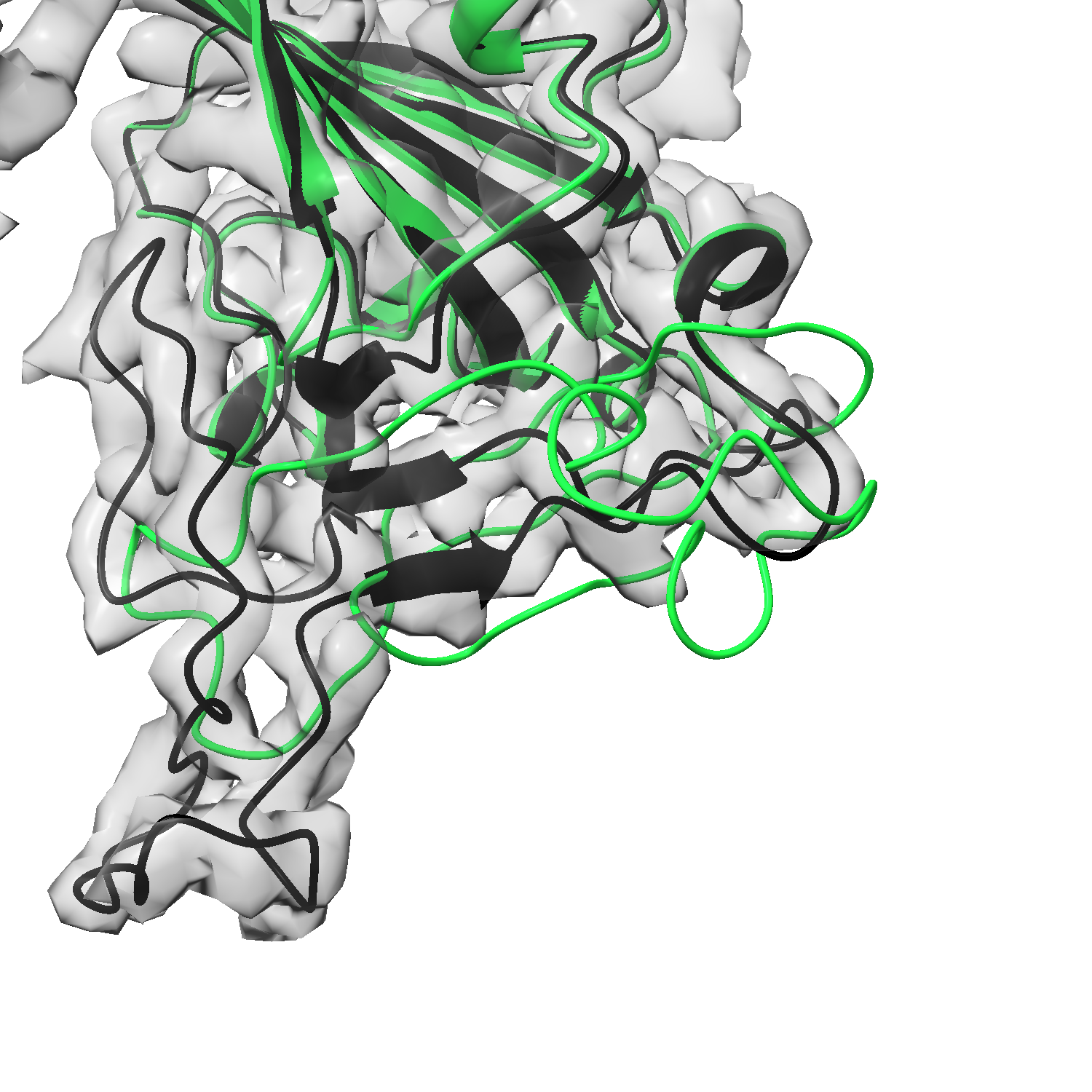}
    \caption{}
    \label{fig:phenix-vs-true}
  \end{subfigure}
  \hspace{0.3cm}
  \begin{subfigure}{0.23\linewidth}
    \includegraphics[scale=0.18]{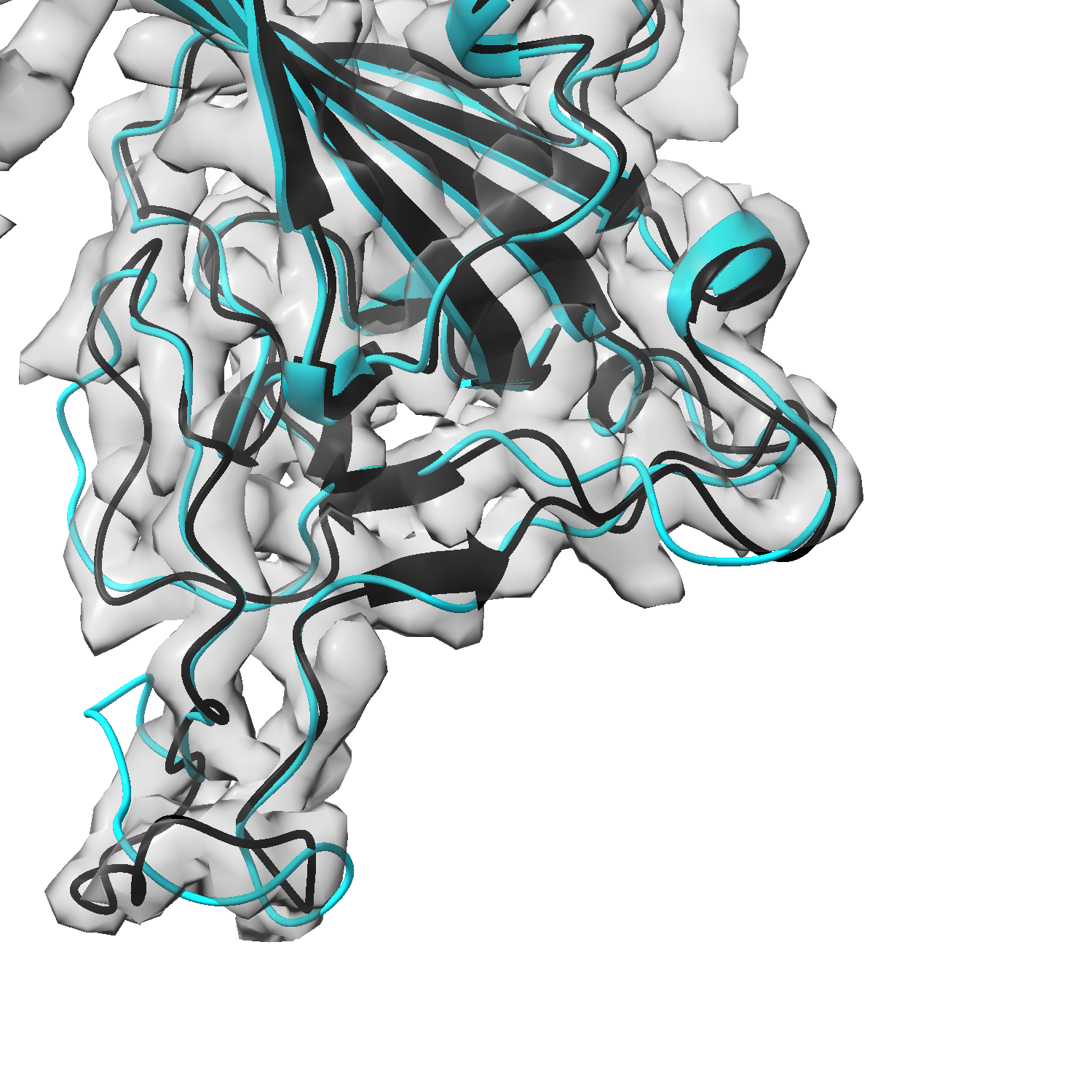}
    \caption{}
    \label{fig:our-vs-true}
  \end{subfigure}
    \caption{Comparison of the deposited structure (black) against results
    from AlphaFold (panel \textbf{a}, TM-score=$0.678$) ,
    traditional flexible fitting (panel \textbf{b}, TM-score=$0.664$),
    implicitly improved method (panel \textbf{c}, TM-score=$0.737$), and
    our method (panel \textbf{d}, TM-score=$0.931$).}
    \label{fig:showcase}
\end{figure}

\begin{table}
  \centering
  \begin{tabular}{@{}lcc@{}}
    \toprule
     & \multicolumn{2}{c}{TM-score}\\
    Method & better group & worse group\\
    \midrule
    AF & 0.960 & 0.751 \\
    AF+MDFF & 0.980 & 0.791 \\
    implicit AF & 0.977 & 0.828\\
    \methname{} & \pmb{0.982} & \pmb{0.852}\\
    \bottomrule
  \end{tabular}
  \caption{Comparison of EM structure prediction quality using four different methods in terms of TM-scores.}
  \label{tab:benchmark}
\end{table}

As a representative example, \cref{fig:showcase} compares the
structure prediction accuracy of all four methods for a
SARS-Cov-2 spike protein (PDB entry: 7M7B, released data: 2021-05-26).
The sequence-based prediction from AlphaFold is far from matching the
target EM map (\cref{fig:showcase}a).
Traditional flexible fitting also failed due to the mismatch of
the initial structure and the target map (\cref{fig:showcase}b).
Implicit AF performed better than these two methods but still
couldn't match the target map exactly (\cref{fig:showcase}c).
With the help of map fragment recognition,
our method \methname{} can more accurately fit the initial structure into the
target map and closely match the deposited structure (\cref{fig:showcase}d).

\subsection{Ablation Study}
The most important part of our method is how to reliably recognize fragments with the given protein sequence and thereby match the predicted structure.

\begin{figure}[!t]
\centering
  \includegraphics[width=0.5\textwidth]{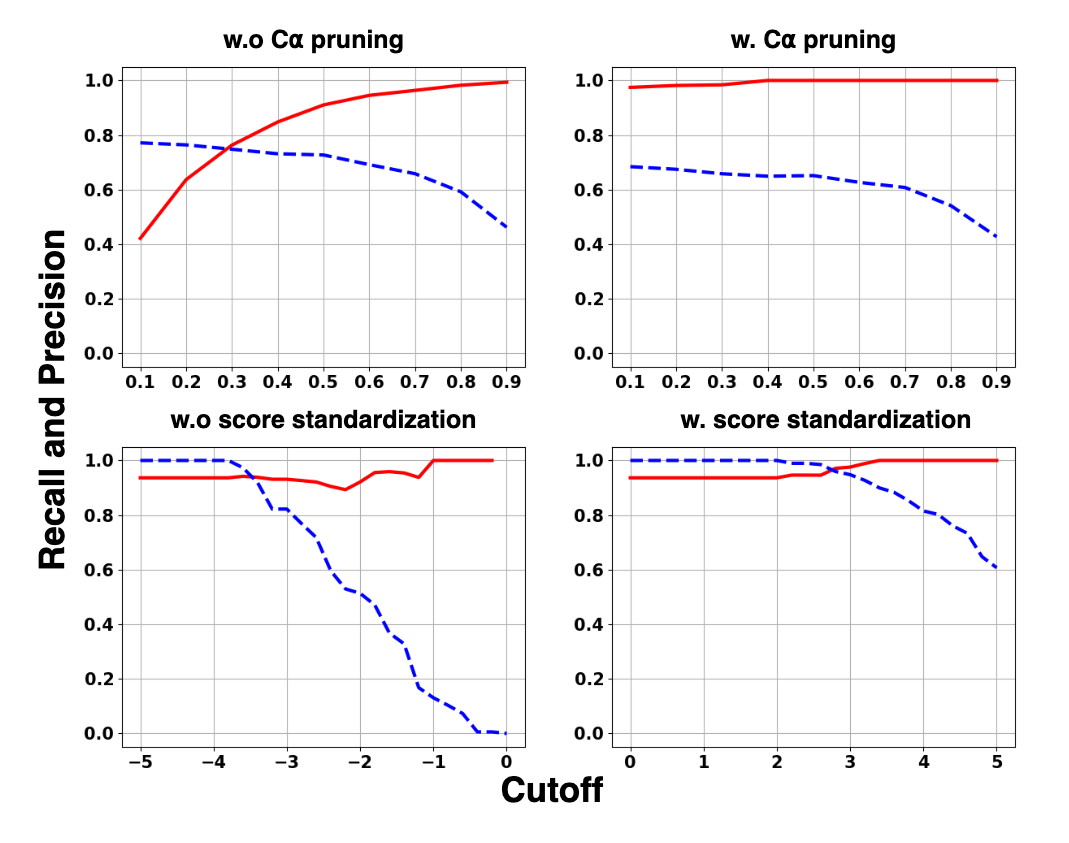}
  \caption{Recall (blue, dashed line) and precision (red, solid line) curves
  as a function of cutoff thresholds (test case: ASCT2).
  The top two panels show the recall/precision vs. \Calpha{}
  detection cutoff values without or with  pruning \Calpha{} fragments shorter than 3.
    The bottom two panels show recall/precision vs. fragment alignment scores
    without or with score standardization.
    }
  \label{fig:ablation}
\end{figure}

Due to the limitation of resolution and model generalization, it's difficult to improve both recall and precision\footnote{True positive \Calpha{} atom is the detected \Calpha{} atom which is within $1.5\angstrom$ of a \Calpha{} atom from the deposited structure. \Calpha{} precision is the fraction of true positive \Calpha{} atoms among all the detected \Calpha{} atoms from the deposited map. \Calpha{} recall is the fraction of true positive \Calpha{} atoms among all the \Calpha{} from the deposited structure.} of \Calpha{} detection. Considering the completeness of protein prediction, our method only focuses on how to exclude false positive \Calpha{} atoms effectively. Increasing the threshold for filtering \Calpha{} atom probability is a simple and direct method, but heavily decreases the recall. Based on the predicted \ppvec{}s, we can directly connect some stable fragments that are usually present in the area with high SNR. In general, the remaining predicted atoms that cannot connect to long fragments usually result from noise, lipids, or some flexible protein loops. Pruning some short fragments or isolated atoms (shorter than 3) can exclude the false positive \Calpha{} atoms and keep the true positive atoms as many as possible with an appropriate threshold (see ~\cref{fig:ablation} top).

\begin{figure}[!t]
\centering
    \includegraphics[scale=0.4]{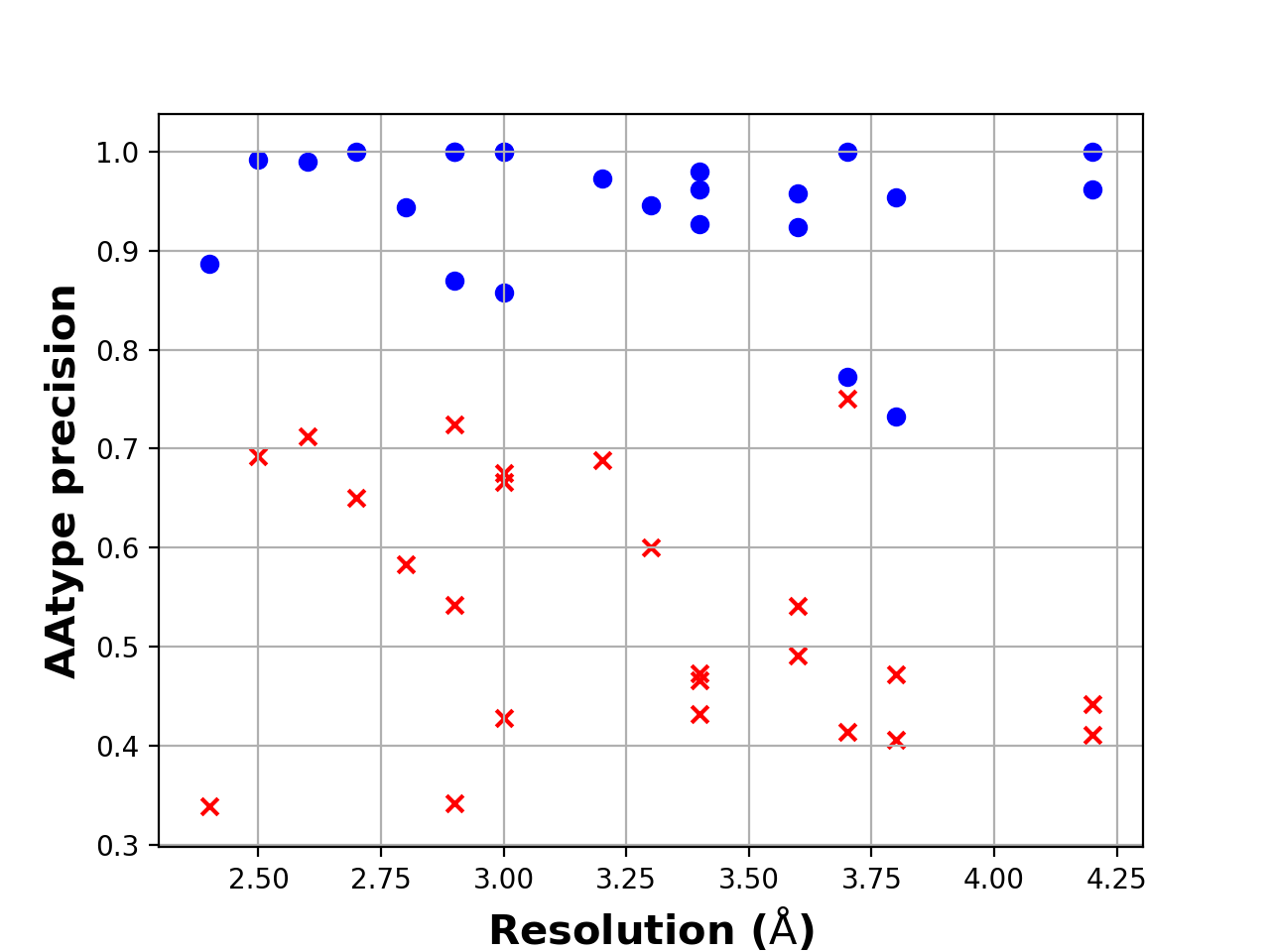}
    \includegraphics[scale=0.4]{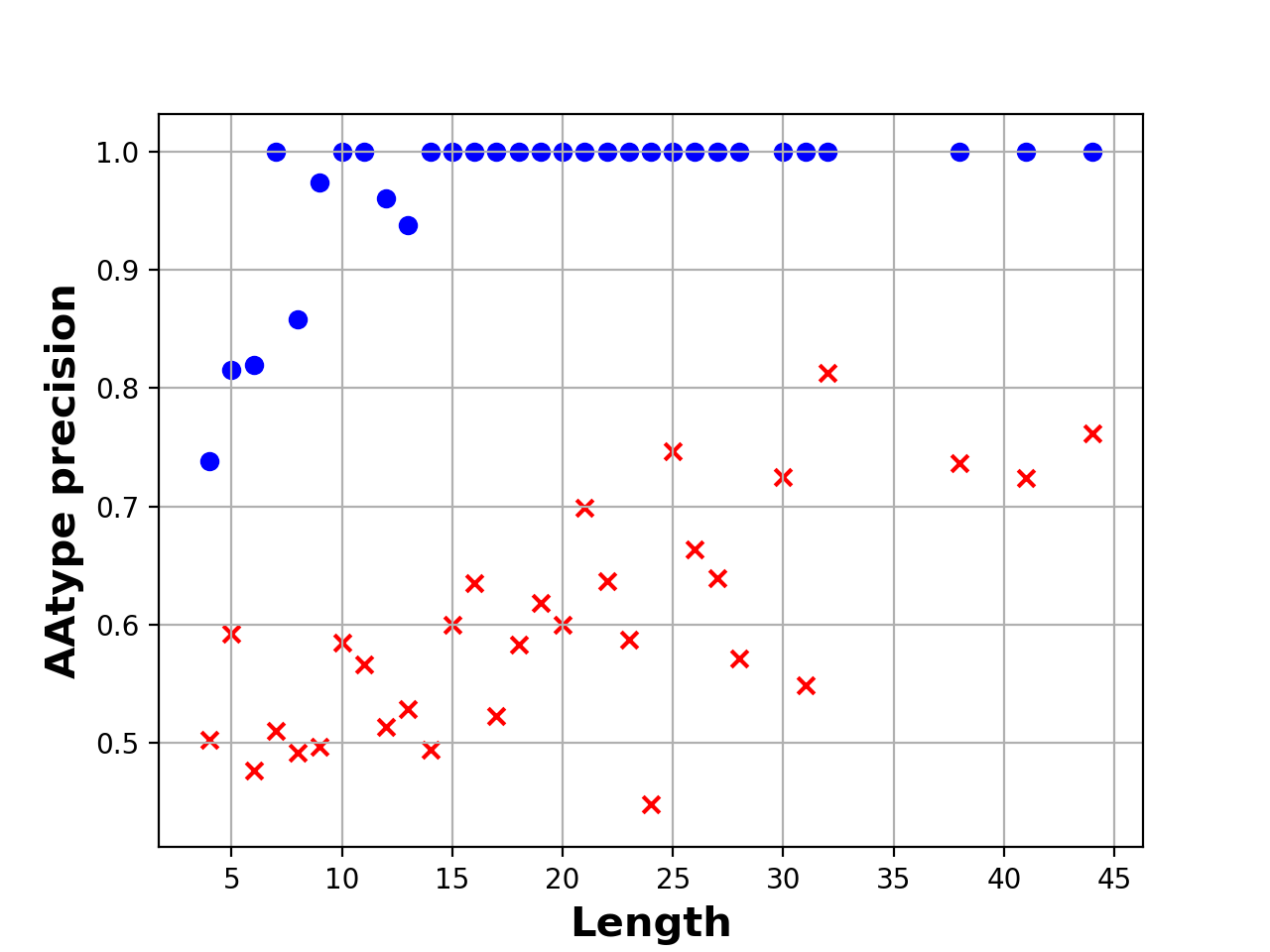}
    \caption{\aatype{} precision between using joint probabilities (blue circle) and individual probabilities (red cross). 
    \textbf{Top}: the \aatype{} precision of two methods among 24 test cases with corresponding map resolution. 
    \textbf{Bottom}: the average \aatype{} precision of two methods for fragments of different length.}
    \label{fig:aatype}
\end{figure}

In previous deep-learning based \textit{de novo} structure building methods~\cite{xu2019a2,pfab2021deeptracer}, only the highest probability of \aatype{} for each residue was used during \aatype{} classification. In our work, it's more reasonable and accurate
to use joint probabilities of consecutive residues to find the optimal alignment with the given
protein sequence. With this method, the average precision of \aatype{} increased from $\sim$50\% to $\sim$80\% in our test cases. To compare the performance between using joint probability of consecutive residues and using the individual probability of each residue,
we evaluated the \aatype{} precisions on 24 experimental cryo-EM maps the same as in section 4.2. All recognized fragments are taken into evaluation without filtering by confidence threshold (see ~\cref{fig:aatype}).

However, for a hypothetical low-resolution map fragment with a nearly uniform probability distribution of \aatype{}s, the optimal alignment chosen by the highest joint probability is of nearly $0\%$ in precision and isn't significantly better than any other alignment for this fragment.
To solve this issue, the confidence of an alignment is measured with score standardization. As results (see ~\cref{fig:ablation} bottom), true \aatype{} alignments can be easily found using a single confidence threshold for all fragments.

%% file: 5-conclusion.tex
\section{Discussion}
\label{sec:Discussion}
In this work, we integrated cryo-EM map recognition
and molecular dynamics simulation techniques into a new method
for cryo-EM structure building.
Our results showed that \methname{} can build structures closely matching
the target map compared to existing methods.
However, there is still room to improve this method.
One of the most important limitations is its sensitivity to map resolution.
From a physical point of view,
resolution is the most important factor that limits the accuracy of C$\alpha$ atom detection and amino acid classification~\cite{si2022artificial}. In practice, the performance of map-recognition-based cryo-EM structure-building methods often degrades when the resolution deteriorates.
Even though our map-recognition network contains coarse-grained recognition modules, the recognition accuracy is still affected for low-resolution maps,
especially for amino acid classification tasks.
Incorrect map-recognition results lead to errors in the protein sequence alignment in the fragment recognition stage and therefore can misguide later structure-fitting steps. Further work is needed to build a complete structure from the low-to-medium-resolution (5--10 \AA{}) cryo-EM maps.

Additionally, other types of macromolecules may present in the cryo-EM maps such as ribonucleic acids (RNA) and/or deoxyribonucleic acid (DNA). Extention to our method is needed to support the structure building of such systems.